\pdfoutput=1

\documentclass[11pt]{article}

\usepackage[preprint]{acl}

\usepackage{times}
\usepackage{latexsym}

\usepackage[T1]{fontenc}

\usepackage[utf8]{inputenc}

\usepackage{microtype}

\usepackage{inconsolata}

\usepackage{graphicx}

\usepackage{amsmath}
\usepackage{lingmacros}
\usepackage{enumitem}
\usepackage{calc}
\usepackage{mathtools}
\usepackage{multirow}
\usepackage{booktabs}
\usepackage{siunitx}

\definecolor{compose}{RGB}{238,127,68}
\definecolor{stack}{RGB}{67,115,181}

\newcommand{\comp}{$\mathrm{-comp}$}

\title{If Attention Serves as a Cognitive Model of Human Memory Retrieval,\\ What is the Plausible Memory Representation?}

\author{
    \textbf{Ryo Yoshida\textsuperscript{\scriptsize $\spadesuit$}}\,
    \textbf{Shinnosuke Isono\textsuperscript{\scriptsize $\spadesuit$\scriptsize $\heartsuit$}}\,
    \textbf{Kohei Kajikawa\textsuperscript{\scriptsize $\spadesuit$\scriptsize $\heartsuit$}}\,
    \textbf{Taiga Someya\textsuperscript{\scriptsize $\spadesuit$}}\,
    \\
    \textbf{Yushi Sugimoto\textsuperscript{\scriptsize $\diamondsuit$}}\,
    \textbf{Yohei Oseki\textsuperscript{\scriptsize $\spadesuit$}}
    \\
    \textsuperscript{\scriptsize $\spadesuit$}The University of Tokyo\,
    \textsuperscript{\scriptsize $\heartsuit$}NINJAL\,
    \textsuperscript{\scriptsize $\diamondsuit$}The University of Osaka
    \\
    \small{
        \texttt{\{yoshiryo0617,taiga98-0809,oseki\}@g.ecc.u-tokyo.ac.jp}
    }
    \\
    \small{
        \texttt{\{s-isono,kohei-kajikawa\}@ninjal.ac.jp}
    }
    \small{
        \texttt{sugimoto.yushi.hmt@osaka-u.ac.jp}
    }
}

\begin{document}
\maketitle

\begin{abstract}
Recent work in computational psycholinguistics has revealed intriguing parallels between attention mechanisms and human memory retrieval, focusing primarily on vanilla Transformers that operate on token-level representations. However, computational psycholinguistic research has also established that syntactic structures provide compelling explanations for human sentence processing that token-level factors cannot fully account for. In this paper, we investigate whether the attention mechanism of Transformer Grammar (TG), which uniquely operates on syntactic structures as representational units, can serve as a cognitive model of human memory retrieval, using Normalized Attention Entropy (NAE) as a linking hypothesis between models and humans. Our experiments demonstrate that TG's attention achieves superior predictive power for self-paced reading times compared to vanilla Transformer's, with further analyses revealing independent contributions from both models. These findings suggest that human sentence processing involves dual memory representations---one based on syntactic structures and another on token sequences---with attention serving as the general memory retrieval algorithm, while highlighting the importance of incorporating syntactic structures as representational units.
\end{abstract}

\section{Introduction}
Whether language models (LMs) developed in natural language processing (NLP) are plausible as cognitive models of human sentence processing is a central question in computational psycholinguistics. Over the past two decades, this question has been primarily addressed from the perspective of \textit{expectation-based theories}---one of the two major classes of human sentence processing theory---examining whether LMs' next-token prediction can serve as a model of human predictive processing~(\citealp{hale2001Probabilistic,levy2008Expectationbased,wilcox2020Predictive,merkx2021Human}; \textit{inter alia}). 

The recent success of Transformers~\cite {vaswani2017Attention} in NLP has unexpectedly opened a new avenue of investigation from the perspective of \textit{memory-based theories}, the other major class of sentence processing theory. Researchers have proposed that the attention mechanism, despite its engineering origins, can implement a human memory retrieval theory known as cue-based retrieval~\cite{vandyke2003Distinguishing}. Recent studies have revealed intriguing parallels between the weighted reference patterns exhibited by the attention mechanism and the elements that humans may retrieve during online sentence comprehension~\citep{ryu2021Accounting,oh2022Entropy,timkey2023Language}.

Computational psycholinguistics has also established that human sentence processing cannot be fully explained by token-level factors; rather, \textit{syntactic structures} have provided compelling explanations for it. For instance, next-token prediction from LMs that explicitly incorporate syntactic structure building demonstrates superior performance in accounting for human brain activity compared to vanilla RNNs and Transformers~\citep{hale2018Finding,wolfman2024Hierarchical}; the number of syntactic nodes hypothesized to be constructed per word correlates significantly with both reading times and neural activity patterns~\citep{kajikawa2024Dissociating,brennan2012Syntactic}.

Given these findings, if attention can serve as a general algorithm for memory retrieval in human sentence processing, human memory retrieval should be captured by the attention mechanism operating on syntactic structures as well as that operating on token sequences. In this paper, we investigate whether the attention mechanism of Transformer Grammar (TG; \citealp{sartran2022Transformera}), which uniquely operates on syntactic structures as representational units, can serve as a cognitive model of human memory retrieval, using Normalized Attention Entropy (NAE; \citealp{oh2022Entropy}) as the linking hypothesis between models and humans. Our experiments demonstrate that TG's attention achieves superior predictive power for self-paced reading times compared to vanilla Transformer's, with further analyses revealing independent contributions from both models. These findings suggest that human sentence processing involves dual memory representations---one based on syntactic structures and another on token sequences---with attention serving as the general memory retrieval algorithm, while highlighting the importance of incorporating syntactic structures as representational units.\footnote{Code for reproducing our results is available at \url{https://github.com/osekilab/TG-NAE}.}


\section{Background}
\label{sec:background}

\subsection{Cue-based retrieval}
\label{subsec:nae}
Many psycholinguistic studies assume that human sentence processing involves memory retrieval, where based on the various cues provided by the current input word (e.g., verbs), elements (e.g., their arguments) are retrieved from working memory. In Example~\ex{1}, taken from \citet{vandyke2002Retrieval}, when the verb \textit{was complaining} is input, its subject \textit{the resident} must be retrieved from working memory.
\eenumsentence[1]{
    \item The worker was surprised that the \textbf{resident}\textsubscript{[subj,anim]} [who was living near the dangerous warehouse] \textit{was complaining} about the investigation.
    \item The worker was surprised that the \textbf{resident}\textsubscript{[subj,anim]} [who said that the warehouse\textsubscript{[subj]} was dangerous] \textit{was complaining} about the investigation.
}
According to the cue-based retrieval theory~\citep{vandyke2003Distinguishing}, such retrieval becomes more difficult when similar elements exist in the sentence because the cues are overloaded; for example, only in Example~\ex{1}b, \textit{warehouse} may interfere with \textit{resident} since they both have the feature [subj] as a retrieval cue. \citet{vandyke2002Retrieval} showed that humans read \textit{was complaining} more slowly in Example~\ex{1}b than in Example~\ex{1}a, providing empirical support for the cue-based retrieval theory. Such interference effects have been observed across various syntactic and semantic features~\citep{vandyke2003Distinguishing,vandyke2011Cuedependent,nicenboim2018Exploratory}.

\subsection{Normalized Attention Entropy (NAE)}
In recent computational psycholinguistics, attempts have been made to interpret the attention mechanism---a weighted reference of preceding tokens based on Query and Key vectors---as a computational implementation of cue-based retrieval. Notably, \citet{ryu2021Accounting} proposed Attention Entropy (AE) as a linking hypothesis, where the diffuseness of attention weights is assumed to quantify the degree of retrieval interference. While AE was initially proposed for modeling interference effects in specific constructions, \citet{oh2022Entropy} extended it to naturally occurring text by introducing two normalizations: (i) division by the maximum entropy achievable given the number of preceding tokens, and (ii) sum-to-1 renormalization of attention weights over preceding tokens  (\textbf{Normalized AE, NAE}).\footnote{\citet{oh2022Entropy} showed that regression models for predicting reading times fail to converge with vanilla AE.}
\begin{align}
\mathrm{NAE}_{l,h,i} &= -\frac{1}{\log_2 |T|} \sum_{j \in T} \tilde{a}_{l,h,i,j} \log_2 \tilde{a}_{l,h,i,j}
\end{align}
where $T$ is the set of preceding token positions,\footnote{For vanilla Transformers, $T=\{1, 2, \ldots, i-1\}$.} $\tilde{a}_{l,h,i,j} = \frac{a_{l,h,i,j}}{\sum_{k \in T} a_{l,h,i,k}}$ is the renormalized attention weight, and $a_{l,h,i,j}$ represents the attention weight from the $i$-th token (Query) to the $j$-th preceding token (Key) in the $h$-th head of layer $l$.\footnote{\citet{oh2022Entropy} explored NAE calculation using various attention weight formulations, but in this paper, we adopt the norm-based attention weight formulation \citep{kobayashi2020Attention}, which achieved the highest predictive power on the self-paced reading time corpus.} In this paper, we employ this NAE as a linking hypothesis between attention mechanisms and human memory retrieval.\footnote{While \citet{oh2022Entropy} also proposed other metrics based on distances between attention weights at consecutive time steps, we exclusively adopt NAE because (i) in TG, the number of preceding elements varies with time, making distance definition non-trivial, and (ii) \citet{oh2022Entropy} demonstrated that NAE's predictive power subsumes that of distance-based metrics in the self-paced reading time corpus.}

\subsection{Transformer Grammar (TG)}
\label{subsec:TG}
\begin{figure}
    \centering
    \includegraphics[width=0.75\linewidth]{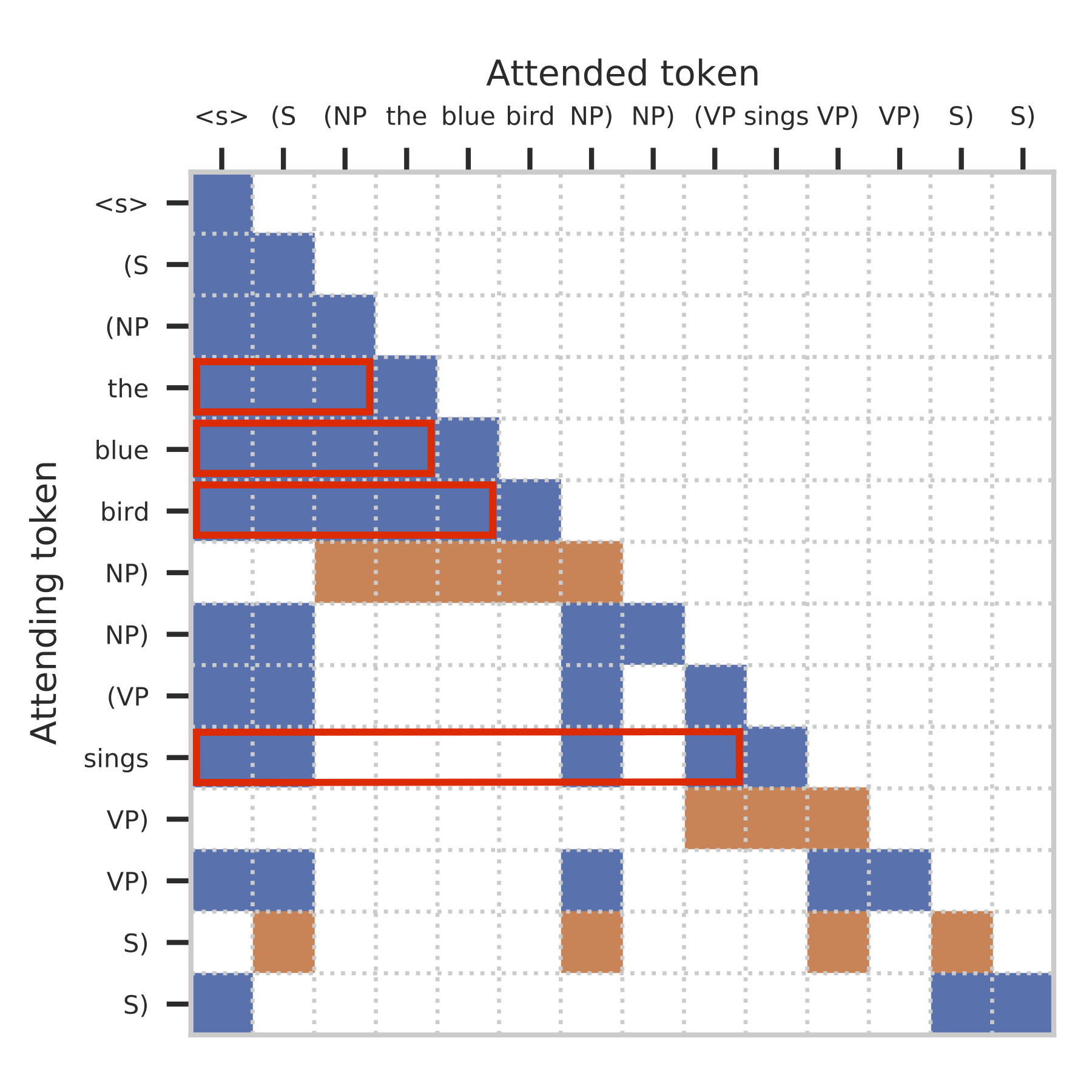}
    \caption{TG's attention mask with \textcolor{compose}{\texttt{COMPOSE}}/\textcolor{stack}{\texttt{STACK}} attention mechanisms, adapted from \citet{sartran2022Transformera}. \textcolor{compose}{\texttt{COMPOSE}} generates a vector representation of the closed phrase, while subsequent \textcolor{stack}{\texttt{STACK}} operations reference this vector as the phrase representation for next-action prediction. Red boxes indicate the attention weights used to calculate NAE for each word.}
    \label{fig:TG}
\end{figure}

\textbf{Transformer Grammar} (\textbf{TG};~\citealp{sartran2022Transformera}) is a type of syntactic LM, a generative model that jointly generates token sequences $\boldsymbol{x}$ and their corresponding syntactic structures $\boldsymbol{y}$. TG formulates the generation of $(\boldsymbol{x}, \boldsymbol{y})$ as modeling a sequence of actions, $\boldsymbol{a}$ (e.g., \texttt{(S (NP The blue bird NP) (VP sings VP) S)}), constructing both token sequences and their syntactic structures in a top-down, left-to-right manner. The action sequence $\boldsymbol{a}$ comprises three types of operations:
\begin{itemize}
    \item \texttt{(X}: Generate a non-terminal symbol \texttt{(X}, where \texttt{X} represents a phrasal tag such as \texttt{NP};
    \item \texttt{w}: Generate a terminal symbol \texttt{w}, where \texttt{w} represents a token such as \texttt{bird};
    \item \texttt{X)}: Generate \texttt{X)} to close the most recent opened non-terminal symbol, where \texttt{X} matches the phrasal tag of the targeted non-terminal symbol.
\end{itemize}
The probability of action sequence $\boldsymbol{a} = (a_1, a_2, \ldots, a_n)$ is decomposed using the chain rule. Formally, TG is defined as:
\begin{align}
    p(\boldsymbol{x}, \boldsymbol{y}) = p(\boldsymbol{a}) = \prod_{t=1}^{n} p(a_t|a_{<t}) .
\end{align}

TG's key innovation lies in its handling of closed phrases: immediately after generating \texttt{X)}, it computes a vector representation of the closed phrase, which subsequent next-action predictions use as the representation for that phrase. Technically, this operation is realized via two components: \texttt{X)} action duplication and a specialized attention mask. The duplication process transforms $\boldsymbol{a}$ into $\boldsymbol{a}'$ by duplicating all \texttt{X)} actions (e.g., \texttt{(S (NP The blue bird NP) NP) (VP sings VP) VP) S) S)}), while preserving the modeling space $p(\boldsymbol{a})$ by preventing predictions for duplicated positions. The attention mask implements two distinct attention mechanisms: \textcolor{compose}{\texttt{COMPOSE}} and \textcolor{stack}{\texttt{STACK}} (Figure~\ref{fig:TG}). \textcolor{compose}{\texttt{COMPOSE}} operates exclusively at the first occurrence of each \texttt{X)} to generate the phrasal representation by attending only to vectors between the corresponding \texttt{(X} and \texttt{X)} (without making predictions). \textcolor{stack}{\texttt{STACK}} operates at all other positions to compute representations for next-action prediction, with attention restricted to positions on the \textit{stack} (comprising unclosed non-terminals, not-composed terminals, and closed phrases).

Previous research has demonstrated that TG's probability estimates align more closely with human offline grammaticality judgments~\citep{sartran2022Transformera} and online brain activity~\citep{wolfman2024Hierarchical} than vanilla Transformers. This paper investigates whether the attention mechanism of TG, which uniquely operates on syntactic structures as representational units, can serve as a cognitive model of human memory retrieval.

\section{Methods}
\label{sec:methods}

\subsection{NAE calculation with TG}
\label{subsec:nae_tg}
The calculation of NAE with TG requires assumptions regarding two key perspectives:
\begin{enumerate}
    \item What syntactic structures should be assumed for a given token sequence?
    \item How should the cognitive load of attention from non-lexical symbols (i.e., \texttt{(X} and \texttt{X)}) be attributed to lexical tokens?
\end{enumerate}

In response to these considerations, we make the following assumptions:
\begin{enumerate}[
    label=\arabic*-A.,
    align=left,
    leftmargin=*,
    labelwidth=\widthof{9.},
    labelsep=0.5em
]
    \item \label{item:oracle} We assume only the globally correct syntactic structure (i.e., ``perfect oracle''; \citealp{brennan2016Naturalistic}).
    \item \label{item:attention} We consider only attention from lexical tokens, excluding attention from non-lexical symbols.
\end{enumerate}

The adoption of \ref{item:oracle} is motivated by two factors. First, the self-paced reading time corpus we utilized here provides gold-standard syntactic structures for each sentence, and previous studies have developed predictors based on these annotations \citep{shain2020FMRI,isono2024Category}. Using the same structural assumptions enables fair comparison with these established predictors, considering the possibility of parsing errors. Second, TG's current implementation lacks beam search procedure~\citep{stern2017Effective,crabbe2019Variable}, an inference technique commonly used in cognitive modeling to handle local ambiguities through parallel parsing~\citep{hale2018Finding,sugimoto2024Localizing}.\footnote{As a proof of concept, we also conducted experiments using multiple syntactic structures generated by word-synchronous beam search with Recurrent Neural Network Grammar \citep{dyer2016Recurrent,kuncoro2017What,noji2021Effective}, obtaining similar results (Appendix~\ref{app:beam}).}

Regarding~\ref{item:attention}, given the multiple possible approaches to attributing processing load from non-lexical symbols to lexical tokens, we adopt the most straightforward and theoretically neutral approach. Figure~\ref{fig:TG} denotes the attention weights used to calculate NAE for each word, with red boxes.

\subsection{Settings}
\paragraph{Language models}
We used 16-layer, 8-head TG and Transformer (252M parameters).\footnote{\url{https://github.com/google-deepmind/transformer_grammars}} All hyperparameters followed the default settings described in \citet{sartran2022Transformera} (see Appendix~\ref{app:hypara}). Following \citet{oh2022Entropy}, we computed NAE separately for each attention head at the top layer and then summed the values across heads.

\paragraph{Training data}
We used BLLIP-\textsc{lg}, a dataset containing 42M tokens (1.8M sentences) from the Brown Laboratory for Linguistic Information Processing (BLLIP) 1987--89 WSJ Corpus Release 1~\citep{charniakeugene2000BLLIP}.\footnote{\url{https://catalog.ldc.upenn.edu/LDC2000T43}} The corpus was re-parsed using a state-of-the-art constituency parser~\citep{kitaev2018Constituency} and split into train-val-test sets by \citet{hu2020Systematic}. BLLIP-\textsc{lg} has been widely used for training syntactic LMs, including TG. Following \citet{sartran2022Transformera}, we trained a 32K SentencePiece tokenizer~\citep{kudo2018SentencePiece} on the training set and segmented each sentence into subword units.

Both TG and Transformer were trained at the sentence level: TG maximized the joint probability $p(\boldsymbol{x}, \boldsymbol{y})$ on action sequences, while Transformer maximized the probability $p(\boldsymbol{x})$ on terminal subword sequences. For training hyperparameters, we largely followed the default settings in \citet{sartran2022Transformera} but adjusted the batch size to fit within the memory constraints of our hardware (NVIDIA A100, 40GB). Accordingly, we tuned other hyperparameters (e.g., learning rate) to maintain training stability.\footnote{For the detailed hyperparameters, see Appendix~\ref{app:hypara}.} We trained three models with different random initialization seeds and selected the checkpoint with the lowest validation loss for each run.

\begin{table*}[]
    \centering
    \begin{tabular}{lllccc}
        \toprule
        Model & $\Delta$LogLik ($\uparrow$) & Predictor & Effect size [\si{\milli\second}] & $p$-value range & Significant seeds \\
        \midrule
        \multirow{2}{*}{TG} & \multirow{2}{*}{\textbf{76.6} ($\pm$8.1)} & \texttt{tg\_nae} & 1.42 ($\pm$ 0.2) & $<$\textbf{0.001} & 3/3 \\
        & & \texttt{tg\_nae\_so} & 2.26 ($\pm$ 0.1) & $<$\textbf{0.001} & 3/3 \\
        \midrule
        \multirow{2}{*}{Transformer} & \multirow{2}{*}{42.8 ($\pm$9.5)} & \texttt{tf\_nae} & 1.32 ($\pm$ 0.2) & $<$\textbf{0.001} & 3/3 \\
        & & \texttt{tf\_nae\_so} & 1.46 ($\pm$ 0.2) & $<$\textbf{0.001} & 3/3 \\
        \bottomrule
    \end{tabular}
    \caption{TG's and Transformer's NAE contribution to reading time prediction ($\Delta$LogLik). The effect size per standard deviation is shown for each model-derived predictor, along with the $p$-value range across random seeds and the number of seeds showing significant contributions. Standard deviations across seeds for $\Delta$LogLik and effect sizes are shown in parentheses. The mean reading time in the analysis is 334~\si{\milli\second}.}
    \label{tab:effect_size}
\end{table*}

\paragraph{Reading time data}
We used the Natural Stories corpus~\citep{futrell2018Natural},\footnote{\url{https://github.com/languageMIT/naturalstories}. We used the corrected version that addresses the data misalignment issue identified in May 2025.} ``a series of English narrative texts designed to contain many low-frequency and psycholinguistically interesting syntactic constructions while still sounding fluent and coherent.'' We selected this corpus for these ``interesting'' syntactic constructions, which provide an ideal testbed for investigating memory retrieval effects that might be less pronounced in simpler, more naturalistic texts. The corpus has also been used in several studies investigating memory-related processing mechanisms~\citep{shain2016Memory,dotlacil2021Parsing,isono2024Category}.

The Natural Stories corpus consists of 10 stories (485 sentences, 10,245 words) with self-paced reading times collected from 181 anonymized native English speakers. Following \citeauthor{futrell2018Natural}'s preprocessing, data points were removed if (i) a participant scored less than 5/6 on comprehension questions for a story or (ii) individual reading times were less than 100~\si{\milli\second} or greater than 3,000~\si{\milli\second}. Following \citet{oh2022Entropy}, we also excluded sentence-initial and sentence-final data points. We further removed sentence-second data points, as they lack the log trigram frequency of the previous token required for our baseline regression model. After preprocessing, 725,875 data points from 180 participants remained for statistical analysis, out of the original 848,875 data points.

\paragraph{Statistical analysis}
We evaluated each LM's NAE contribution to reading time prediction by measuring improvements in regression model fit when adding NAE as predictors. For each model (TG/Transformer), we included both the current word's NAE (\texttt{tg\_nae}/\texttt{tf\_nae}) and the previous word's NAE (\texttt{tg\_nae\_so}/\texttt{tf\_nae\_so}) to account for spillover effects~\citep{mitchell1984Evaluation}.\footnote{\texttt{\_so} indicates spillover.}\footnote{Following~\citet{oh2022Entropy}, we summed the subword NAE values for each word.} Model improvement was quantified as the increase in log-likelihood ($\Delta$LogLik). This evaluation was conducted for each random seed, and we report the mean $\Delta$LogLik with standard deviation.

Following previous studies such as~\citet{dotlacil2021Parsing},~\citet{shain2016Memory}, and~\citet{isono2024Category}, the baseline regression model controlled for non-structural, basic aspects of text known to affect reading times:
\begin{itemize}
    \item \texttt{zone} and \texttt{position} (integer): word position in the story and sentence;
    \item \texttt{wordlen} (integer): number of characters in the word;
    \item \texttt{unigram}, \texttt{bigram}, and \texttt{trigram} (continuous): log-transformed n-gram frequencies.
\end{itemize}
We additionally included the following predictors:
\begin{itemize}
    \item \texttt{tg\_surp} and \texttt{tf\_surp} (continuous): surprisal from TG and Transformer;
    \item \texttt{stack\_count} (integer): number of elements in the \textit{stack} (comprising unclosed non-terminals, not-composed terminals, and closed phrases).
\end{itemize}
Following \citet{oh2022Entropy}, we included surprisal to test NAE's significance in the presence of surprisal predictors from the same LMs.\footnote{For an experiment on the predictive power of surprisal itself, see Appendix~\ref{app:surp}.} Stack count was included to isolate the cost of holding elements~\cite{joshi1990Processing,abney1991Memory,resnik1992LeftCorner} from their interference effects, which TG's NAE was designed to capture. For the correlations between the predictors, see Appendix~\ref{app:corr}. 

All predictors were \textit{z}-transformed, and we also included the previous word's values as predictors to model spillover, except for the positional information. The baseline regression model was a linear mixed-effects 
model~\citep{baayen2008Mixedeffects} with these fixed effects and by-subject and by-story random intercepts:
\begin{align}
    \label{eq:regression}
    \log(\texttt{RT}) \sim{}
        & \texttt{zone} + \texttt{position} + \texttt{wordlen}~+\notag \\
        & \texttt{unigram} + \texttt{bigram} + \texttt{trigram}~+\notag \\
        & \texttt{tg\_surp} + \texttt{tf\_surp}~+ \notag \\
        & \texttt{stack\_count} + \texttt{wordlen\_so}~+ \notag \\
        & \texttt{unigram\_so} + \texttt{bigram\_so}~+ \notag \\
        & \texttt{trigram\_so} + \texttt{tg\_surp\_so}~+ \notag \\
        & \texttt{tf\_surp\_so} + \texttt{stack\_count\_so}~+ \notag \\
        & (1 \,|\, \texttt{participant}) + (1 \,|\, \texttt{story})
\end{align}

To assess each LM's independent contribution to reading time prediction, we also conducted likelihood ratio tests~\citep{wurm2014What} by extending Equation~\ref{eq:regression} in two ways: adding both LMs' NAE versus adding only one LM's NAE. Note that a larger $\Delta$LogLik from one LM does not necessarily indicate that it contributes above and beyond the other LM, nor does a smaller $\Delta$LogLik indicate no unique contribution. Following \citet{aurnhammer2019Comparing}, we used NAE and surprisal values averaged across random seeds for these nested model comparisons.

\section{Results}
\label{sec:results}

\subsection{Does TG's NAE have predictive power for reading times?}
\label{subsec:loglik}
Table~\ref{tab:effect_size} presents the contributions of TG's and Transformer's NAE to reading time prediction. First, Transformer's NAE exhibited significant predictive power for reading times, independent of baseline predictors such as surprisal. The effect size was in the expected positive direction (higher NAE values corresponding to longer reading times), showing both immediate and spillover effects. This corroborates the arguments of \citet{ryu2021Accounting} and \citet{oh2022Entropy} that the attention mechanism---the weighted reference of preceding tokens---functions as a cognitive model of human memory retrieval, despite its engineering-oriented origins.

Second, TG's NAE exhibited robust predictive power, demonstrating significant positive effects in both immediate and spillover conditions. This finding not only provides additional evidence for incremental construction of syntactic structures in human sentence processing (e.g.,~\citealp{fossum2012Sequential}), but also suggests that TG's attention mechanism effectively models memory retrieval from these constructed syntactic representations.

Finally, TG's NAE made a substantially stronger contribution to reading time prediction ($\Delta$LogLik$=$76.6) compared to Transformer's NAE ($\Delta$LogLik$=$42.8). This finding suggests that retrieval from syntactic memory representations plays a more dominant role in human sentence processing than retrieval from lexical memory representations. This underscores the importance of incorporating syntactic structures as a unit of memory representation, which we implemented through the integration of TG and NAE here.

\subsection{Do TG's and Transformer's NAE have independent contributions?}
\label{subsec:nested}
\begin{figure}
    \centering
    \includegraphics[width=\linewidth]{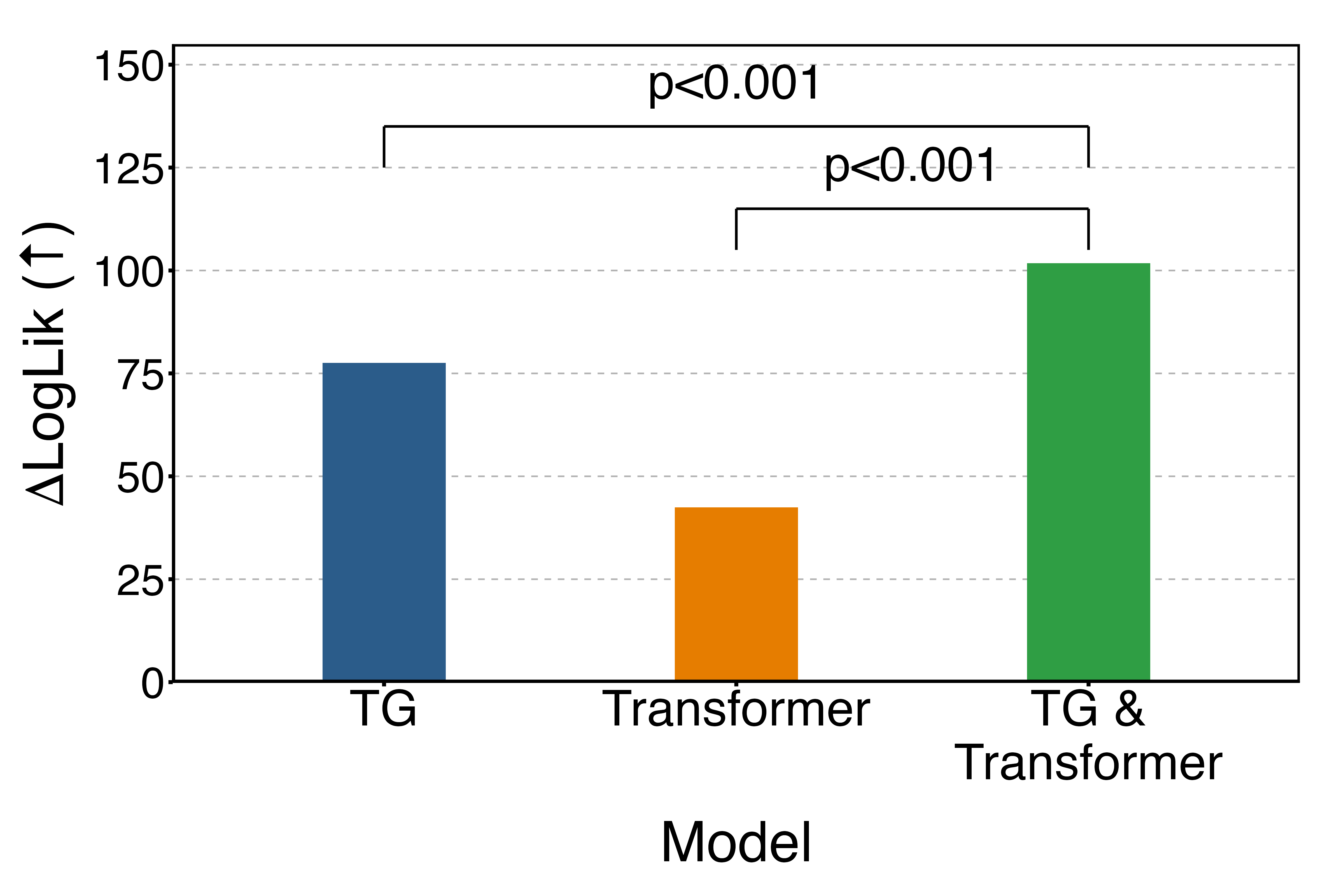}
    \caption{Likelihood ratio test results examining the independence of NAE's predictive power}
    \label{fig:nested}
\end{figure}

Figure~\ref{fig:nested} presents the results of likelihood ratio tests examining the independence of TG's and Transformer's NAE contributions. The regression model incorporating NAE from both LMs (`TG~\&~Transformer') demonstrated significantly higher predictive power than the models containing NAE from either LM alone (`TG' or `Transformer'). This reveals that TG's NAE certainly captures variance in reading times that Transformer's NAE cannot explain, while Transformer's NAE, despite its lower overall predictive power, accounts for unique variance not captured by TG's NAE. This finding aligns with psycholinguistic literature, where cognitive models of memory retrieval encompass both syntax-based approaches (e.g., verb-argument relationships; \citealp{lewis2005Activationbased}) and semantic-based approaches (e.g., bag-of-words-like similarity; \citealp{brouwer2012Getting}), suggesting that the attention mechanisms of TG and Transformer serve as complementary cognitive models, each capturing distinct aspects of human memory retrieval.

\subsection{What aspects of memory retrieval do TG's and Transformer's NAE capture?}
\label{subsec:pos}
\begin{figure*}
    \centering
    \includegraphics[width=\linewidth]{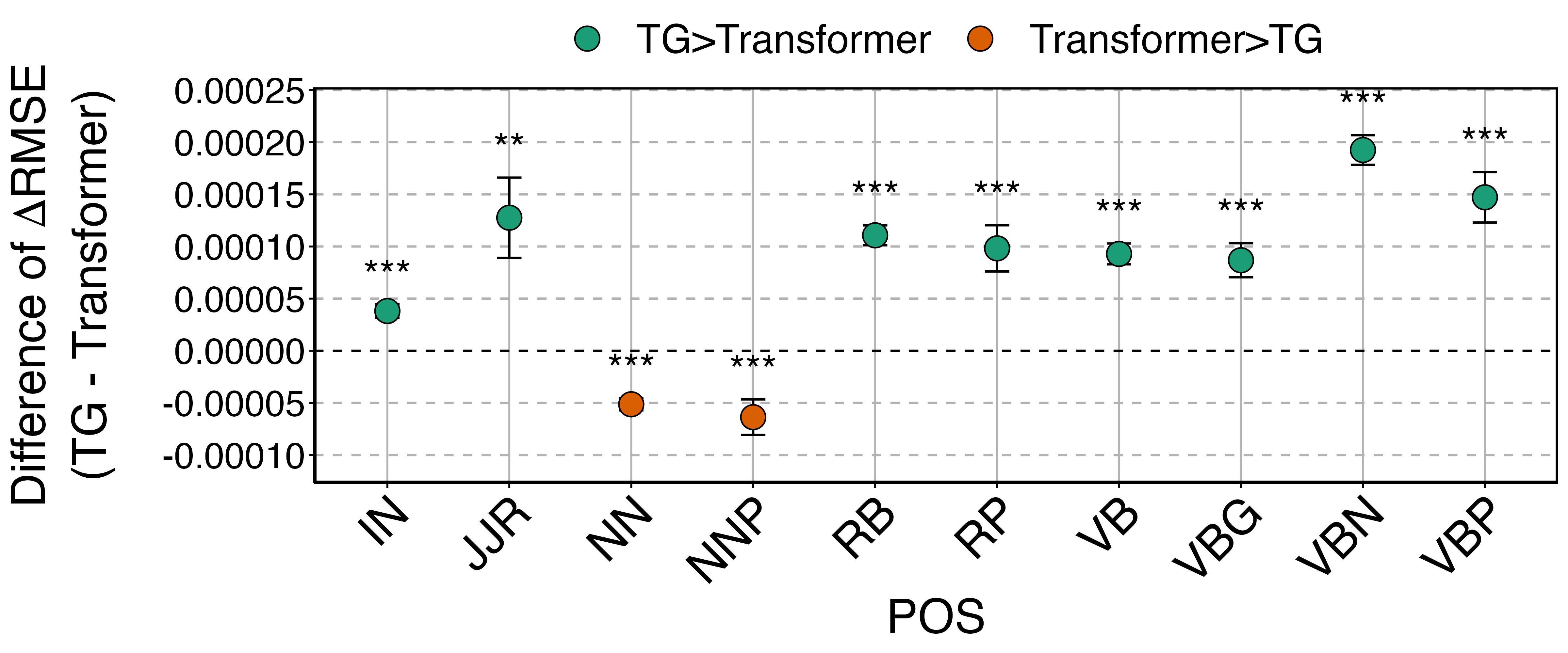}
    \caption{Differences in reading time prediction improvement ($\Delta$RMSE) between TG and Transformer across POS tags (TG~-~Transformer). The y-axis shows the mean differences per word, with the error bars representing standard errors. Only POS tags showing significant improvement in either model and significant differences between models are displayed. Statistical significance after Bonferroni correction: ** $p<0.01$, *** $p<0.001$.}
    \label{fig:pos}
\end{figure*}

To investigate the aspects of human memory retrieval captured by TG's and Transformer's NAE, we analyzed differences in prediction improvement across part-of-speech (POS) tags annotated in the Natural Stories corpus.\footnote{For a complete list of POS tags in the Natural Stories corpus, see Appendix~\ref{app:pos}.} Our analysis followed three steps: (i) selecting POS tags with more than 1,000 occurrences, (ii) for each POS tag, testing the significance of improvement from the baseline regression model (measured in $\Delta$ Root Mean Squared Error, $\Delta$RMSE) when adding NAE of the current and previous word as fixed effects,\footnote{We used the same regression models as in Section~\ref{subsec:nested}, where surprisal and NAE values were averaged across seeds.} and (iii) examining the significance of \textit{differences} in $\Delta$RMSE between TG and Transformer for POS tags where either model showed significant improvement. We assessed significance using Wilcoxon signed-rank tests with Bonferroni correction ($p<0.05$).

Figure~\ref{fig:pos} presents the differences in prediction improvement across POS tags. Consistent with the larger $\Delta$LogLik value, TG's NAE demonstrated advantages over Transformer's NAE across a broader range of POS tags. Notably, TG's NAE exhibited superior improvement across verbs (\texttt{VB}, \texttt{VBG}, \texttt{VBN}, and \texttt{VBP}), while Transformer's NAE excelled for nouns (\texttt{NN} and \texttt{NNP}). This pattern aligns with our earlier argument regarding the complementary nature of these models (Section~\ref{subsec:nested}), indicating that different types of retrieval operations---verb-triggered retrieval, which often relies on syntactic features (e.g., argument structure), and noun-triggered retrieval, which often relies on semantic features (e.g., referential associations)---are better captured by distinct attention mechanisms: attention with syntactic and token memory representations, respectively.

\section{Follow-up analysis}
\label{sec:analysis}

\subsection{Do TG's advantages stem from the COMPOSE attention?}
\label{subsec:compose}
\begin{table}[]
    \centering
    \newlength{\modelwidth}
    \newlength{\loglikwidth}
    \newlength{\predictorwidth}
    \newlength{\pvaluewidth}
    \newlength{\starswidth}
    \settowidth{\starswidth}{***}
    \settowidth{\modelwidth}{TG\textsubscript{\comp}}
    \settowidth{\loglikwidth}{$\Delta$LogLik}
    \settowidth{\predictorwidth}{\texttt{*\_nae\_so}}
    \settowidth{\pvaluewidth}{Sig. seeds}
    
    \begin{tabular}{@{}
        p{\modelwidth}
        p{\loglikwidth}
        p{\predictorwidth}
        p{\pvaluewidth}
        @{}}
        \toprule
         Model & $\Delta$LogLik & Predictor & $p$-value \\
         \midrule
         \multirow{2}{*}{TG} & \textbf{46.1} & \texttt{*\_nae} & \makebox[\starswidth][l]{$^{**}$} (2/3) \\
         & ($\pm$ 9.1) & \texttt{*\_nae\_so} & \makebox[\starswidth][l]{$^{***}$} (3/3) \\
         \midrule
         \multirow{2}{*}{TG\textsubscript{\comp}} & 18.1 & \texttt{*\_nae} & \makebox[\starswidth][l]{$^{**}$} (1/3) \\
         & ($\pm$ 9.3) & \texttt{*\_nae\_so} & \makebox[\starswidth][l]{$^{***}$} (3/3) \\
        \bottomrule
    \end{tabular}
    \caption{TG's and TG\textsubscript{\comp}'s contribution to reading time prediction. The rightmost column shows the $p$-value range across random seeds that achieved significance ($^{***}$ $p < 0.001$ and $^{**}$ $p < 0.01$), along with the number of seeds showing significant contributions. Due to the potential multicollinearity between the Transformer's NAE and TG/TG\textsubscript{\comp}'s NAE, the column of the effect size is omitted.}
    \label{tab:nested_txlTree}
\end{table}

\begin{figure}
    \centering
    \includegraphics[width=\linewidth]{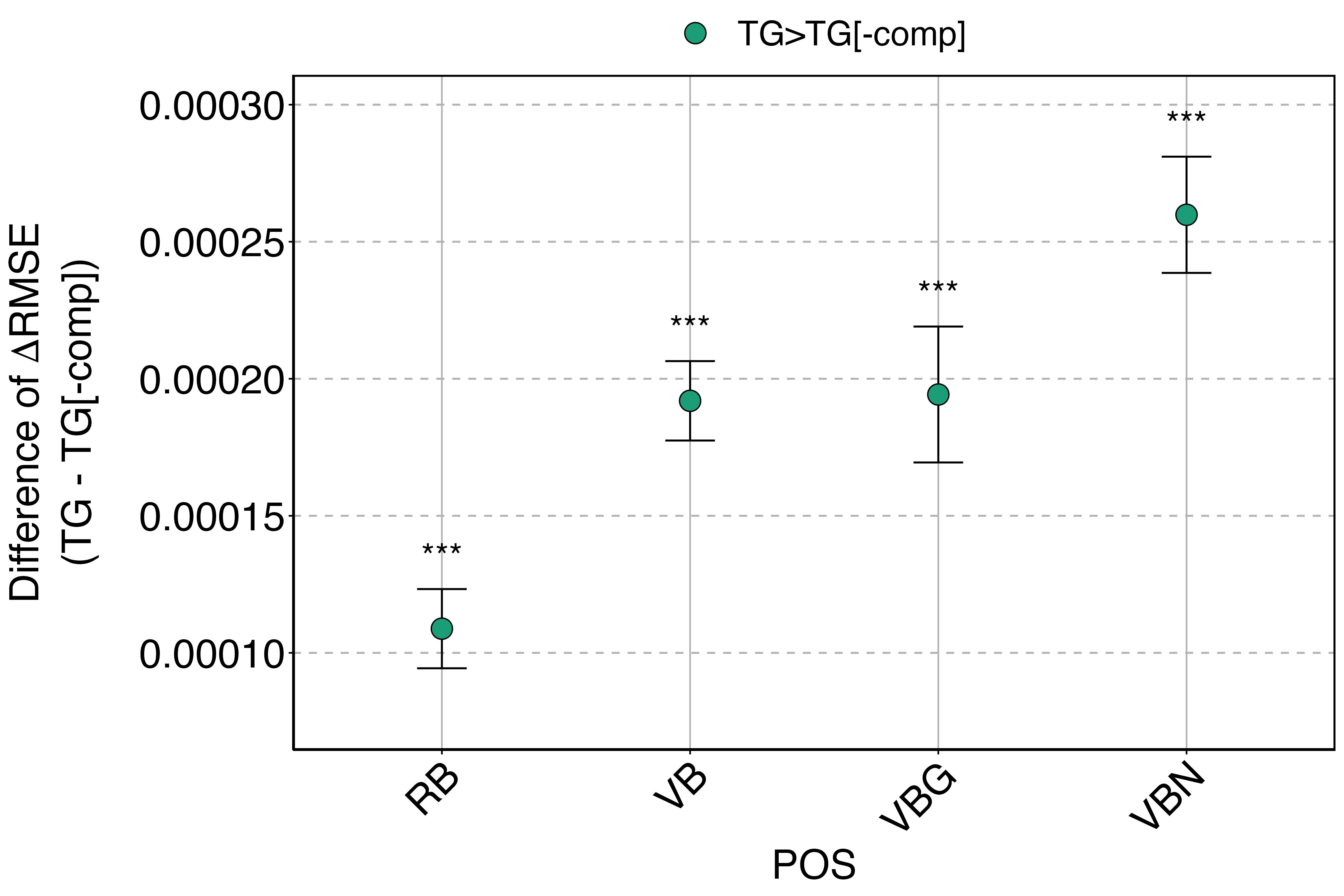}
    \caption{Differences in $\Delta$RMSE between TG and TG\textsubscript{\comp} across POS tags (TG - TG\textsubscript{\comp}). Statistical significance after Bonferroni correction: *** $p<0.001$.}
    \label{fig:pos_txlTree}
\end{figure}

As described in Section~\ref{subsec:TG}, TG's key feature is the COMPOSE attention, which explicitly generates single vector representations for closed phrases. Here, we investigate whether TG's predictive power derives from merely considering syntactic structures or from explicitly treating closed phrases as single representations (see ~\citealp{hale2018Finding,brennan2020Localizing}). To address this question, we developed TG\textsubscript{\comp}, a TG variant that processes each action in the action sequence $\boldsymbol{a}$ as an individual token without the COMPOSE attention (i.e., \citeauthor{choe2016Parsing}'s \textit{Parsing as Language Modeling} approach). We trained TG\textsubscript{\comp} with identical hyperparameters as TG. The baseline regression model (Equation~\ref{eq:regression}) was augmented with (i) TG\textsubscript{\comp}'s surprisal and (ii) Transformer's NAE to (i) ensure a fair comparison between TG and TG\textsubscript{\comp} and (ii) distinguish between the effects of direct terminal token access and syntactic structure consideration in TG\textsubscript{\comp}.\footnote{For direct comparison between TG and TG\textsubscript{\comp} under the baseline regression model without Transformer's NAE, see Appendix~\ref{app:weak}.}

Table~\ref{tab:nested_txlTree} presents the $\Delta$LogLik values obtained when incorporating either TG's or TG\textsubscript{\comp}'s NAE as fixed effects into the baseline regression model. Note that due to the potential multicollinearity between Transformer's NAE and TG/TG\textsubscript{\comp}'s NAE, we focus on the $\Delta$LogLik values and significance of the contribution rather than individual effect sizes. 

Our analysis reveals two key findings. First, TG\textsubscript{\comp}'s NAE demonstrates significant predictive power for reading times, even in the presence of Transformer's NAE, implying that consideration of syntactic structures alone captures certain memory retrievals that the token-level attention mechanism cannot capture. Second, TG's NAE outperforms TG\textsubscript{\comp}'s, suggesting that the attention mechanism that treats closed phrases as single representations more effectively captures variance in syntax-based memory retrieval. The likelihood ratio tests further revealed that TG's NAE captured reading time patterns unexplainable by TG\textsubscript{\comp} (`TG~\&~TG\textsubscript{\comp}'$>$`TG\textsubscript{\comp}', $p<0.001$), while TG\textsubscript{\comp} did not explain unique variance beyond what TG already accounts for (`TG~\&~TG\textsubscript{\comp}'$>$`TG', $p=0.478$).

We analyzed the $\Delta$RMSE differences across POS tags to investigate which aspects of human memory retrieval are better captured by the COMPOSE attention. Figure~\ref{fig:pos_txlTree} showed that TG's NAE was consistently superior to TG\textsubscript{\comp}'s NAE across verbs (\texttt{VB}, \texttt{VBG}, and \texttt{VBN}), highlighting the critical role of the COMPOSE attention in capturing verb-triggered retrieval, a type of retrieval that was identified as distinctively better captured by TG compared to vanilla Transformers (Section~\ref{subsec:pos}).

\subsection{Does TG's NAE capture interference effects?}

Psycholinguistic research has considered two primary types of memory retrieval costs: \textit{interference} effects, which NAE aims to capture, and \textit{decay} effects---the cognitive load associated with accessing elements at greater linear distances (e.g.,~\citealp{gibson1998Linguistic,gibson2000Dependency}). Here, we examine whether TG's NAE genuinely captures interference effects by testing its independence from variables that model memory decay effects. For modeling decay effects, we employed Category Locality Theory (CLT;~\citealp{isono2024Category}),\footnote{Although Dependency Locality Theory (DLT;~\citealp{gibson1998Linguistic,gibson2000Dependency}) is widely recognized as one of the most prominent models for capturing decay effects, we opted for CLT here, following \citeauthor{isono2024Category}'s finding that DLT-based predictors fail to achieve statistical significance in explaining reading times in the Natural Stories corpus.} which treats phrases in syntactic structure\footnote{CLT assumes syntactic structure based on Combinatory Categorial Grammar~\citep{steedman2000Syntactic}.} as representational units of memory and quantifies decay effects using the distance (measured in content words) between an input and the phrases to be composed with it.

To assess independence, we tested whether TG's NAE and CLT maintain their contributions when simultaneously included in the baseline regression model (Equation~\ref{eq:regression}), and examined their independence through likelihood ratio tests.\footnote{As in other likelihood ratio tests, we used surprisal and NAE values averaged across random seeds.} The results (Table~\ref{tab:clt}) show that TG's NAE exhibited significant effects in both immediate and spillover conditions, and CLT demonstrated a significant spillover effect. The likelihood ratio tests confirmed that these effects were independent (`TG~\&~CLT'$>$`CLT', $p<0.001$; `TG~\&~CLT'$>$`TG', $p<0.001$).

These results provide empirical evidence that NAE quantifies interference rather than decay in memory retrieval---extending beyond previous studies on NAE~\citep{ryu2021Accounting,oh2022Entropy}. This finding is significant because, as far as we are aware, while psycholinguistics has developed various implementations of memory decay effects, it has lacked broad-coverage implementations of interference effects applicable to naturally occurring texts. Our results suggest that NAE represents a promising approach for quantifying interference effects in a broad-coverage manner.

\begin{table}[]
    \centering
    \begin{tabular}{l l S[table-format=1.2]}
        \toprule
        Model & Predictor & {Effect size [\si{\milli\second}]} \\
        \midrule
        \multirow{4}{*}{TG \& CLT} 
            & \texttt{tg\_nae}    & 1.18$^{***}$ \\
            & \texttt{tg\_nae\_so} & 2.38$^{***}$ \\
            \cmidrule(lr){2-3}
            & \texttt{clt}        & 0.06$^{\phantom{***}}$ \\
            & \texttt{clt\_so}     & 1.30$^{***}$ \\
        \bottomrule
    \end{tabular}
    \caption{Effect sizes per standard deviation are shown for TG's NAE and CLT predictors. Significance levels: $^{***}$ $p < 0.001$.}
    \label{tab:clt}
\end{table}

\section{Level of description}
\label{sec:marr}

In cognitive modeling studies based on surprisal theory, explanations typically follow the form ``if these LMs were models of human prediction, the difficulty of next-word disambiguation that humans solve would be approximated as follows.'' Such explanations typically operate at the most abstract of \citeauthor{marr1982vision}'s three levels of description---the computational level. Recently, \citet{futrell2020LossyContext} proposed lossy-context surprisal to integrate memory representation perspectives into surprisal theory. However, as they explicitly stated, this theory remains at the computational level, relaxing assumptions about memory representations in human predictive processing. In contrast, cognitive models of human memory, such as cue-based retrieval, generally provide explanations about \textit{mechanisms} that deal with specific mental representations. These explanations typically move down one level to the algorithmic level of description. 

While not explicitly stated by the authors, we argue that the work of \citet{ryu2021Accounting} and \citet{oh2022Entropy}---conceptualizing attention mechanisms as implementations of cue-based retrieval---represents movement toward the algorithmic level, approaching cue-based retrieval itself. Our research advances this direction by investigating a fundamental question at this level: the nature of memory representations (see~\citealp{hale2014automaton}). Future work could fully operate at the algorithmic level by incorporating more realistic elements such as parallel parsing, left-corner parsing strategies, and memory decay mechanisms.

\section{Conclusion}
In this paper, we have demonstrated that attention can serve as the general algorithm for modeling human memory retrieval from two representational systems. Furthermore, we have shown that among the LMs examined in this paper (TG, TG\textsubscript{\comp}, and Transformer), TG---whose attention mechanism uniquely operates on syntactic structures as representational units---best captures dominant factors in human sentence processing. Our results suggest that the integration of attention mechanisms (developed in NLP) with syntactic structures (theorized in linguistics) constitutes a broad-coverage candidate implementation for human memory retrieval. We hope these findings will foster greater collaboration between these two fields.

\section*{Limitations}
Our NAE calculation comprised three steps: (i) computing NAE for each attention head in the top layer, (ii) adding the values across heads, and (iii) summing subword-level values into word level. While this procedure strictly adhered to ~\citet{oh2022Entropy}, alternative approaches to handling layers, attention heads~\citep{ryu2021Accounting}, and subword tokens~\citep{oh2024Leading,giulianelli2024Proper} warrant investigation.

While our study provides an in-depth investigation using the Natural Stories corpus---an English self-paced reading time corpus containing many interesting syntactic constructions---the breadth of our analysis has certain limitations. The generalizability of our findings to different languages (e.g., Japanese self-paced reading time corpus from~\citealp{asahara2022Reading}) and other cognitive load (e.g., gaze duration from~\citealp{kennedy2003dundee} or EEG and fMRI from~\citealp{bhattasali2020Alice}) remains to be investigated.

As discussed in Section~\ref{subsec:nae_tg}, we employed ``perfect oracles'' as syntactic structures behind token sequences. While this assumption has been widely adopted in previous studies on human syntactic processing~\citep{brennan2016Naturalistic,shain2016Memory,stanojevic2021Modeling,isono2024Category}, this idealization leaves the resolution of local ambiguities, which humans encounter during actual sentence processing, outside the scope of our study (for a conceptual case study, see Appendix~\ref{app:beam}).

Although we adopted the default TG implementation of a top-down parsing strategy, psycholinguistic literature has suggested that a left-corner parsing strategy might be more plausible for human sentence processing from a perspective of memory capacity (cf. \texttt{stack\_count})~\citep{abney1991Memory,resnik1992LeftCorner}. However, when contrasting memory representations based on syntactic structures versus token sequences, the attention mechanism of the top-down TG can serve as a reasonable approximation of retrieval from structure-based memory---this aligns with previous work contrasting predictions based on syntactic structures versus token sequences, which used top-down TG or RNNG to represent structure-based prediction~~\citep{wolfman2024Hierarchical,hale2018Finding,brennan2020Localizing}.

Finally, while this paper focused on investigating the attention mechanism through the lens of memory-based theory, exploring TG and Transformer as integrated implementations for expectation-based theory (via surprisal) and memory-based theory (via NAE) represents a promising future direction~\citep{michaelov2021Different,ryu2022using}. Specifically, future work could investigate the attention mechanism as the underlying driver of surprisal's predictive power (Appendix~\ref{app:surp}), analyzing the relationship between surprisal and NAE.

\section*{Ethical considerations}
We employed AI-based tools (Claude, ChatGPT, GitHub Copilot, and Grammarly) for writing and coding assistance. These tools were used in compliance with the ACL Policy on the Use of AI Writing Assistance.

\section*{Acknowledgments}
We sincerely thank Michael Wolfman and John Hale for providing detailed information on \citet{wolfman2024Hierarchical}. We also appreciate the insightful reviews provided by anonymous ARR 2025 February reviewers. This work was supported by JSPS KAKENHI Grant Number 24H00087, Grant-in-Aid for JSPS Fellows JP24KJ0800, JST PRESTO Grant Number JPMJPR21C2, and JST SPRING Grant Number JPMJSP2108.

\bibliography{anthology,custom}

\clearpage
\appendix
\section{Hyperparameters}
\label{app:hypara}
\begin{table*}
    \centering
    \begin{tabular}{lll}
        \toprule
         Model architecture & Transformer-XL~\citep{dai2019TransformerXL} \\
         Vocabulary size & 32,768 \\
         Model dimension & 1,024 \\
         Feed-forward dimension & 4,096 \\
         Number of layers & 16 \\
         Number of heads & 8 \\
         Segment length & 256 \\
         Memory length & 256 \\
         \midrule
         Optimizer & Adam ($\beta_1=0.9$, $\beta_2=0.999$) ~\citep{DBLP:journals/corr/KingmaB14} \\
         Batch size & 16 \\
         Number of training steps & 400,000 \\
         Learning rate scheduler & Linear warm-up \& cosine annealing \\
         Number of warm-up steps & 32,000 \\
         Initial learning rate & $2.5\times10^{-8}$ \\
         Maximum learning rate & $3.75 \times 10^{-5}$ \\
         Final learning rate & $7.5\times10^{-8}$ \\
         Dropout rate & 0.1 \\
        \bottomrule
    \end{tabular}
    \caption{Model and training hyperparameters}
    \label{tab:hypara}
\end{table*}

The hyperparameters are shown in Table~\ref{tab:hypara}. All model hyperparameters follow~\citet{sartran2022Transformera,wolfman2024Hierarchical}, while training hyperparameters were adjusted to accommodate the batch size suitable for our computational resources (NVIDIA A100, 40GB). The total computational cost required for all experiments was approximately 225 GPU hours.

\section{Correlations between the predictors}
\label{app:corr}

\begin{figure*}
    \centering
    \includegraphics[width=0.80\linewidth]{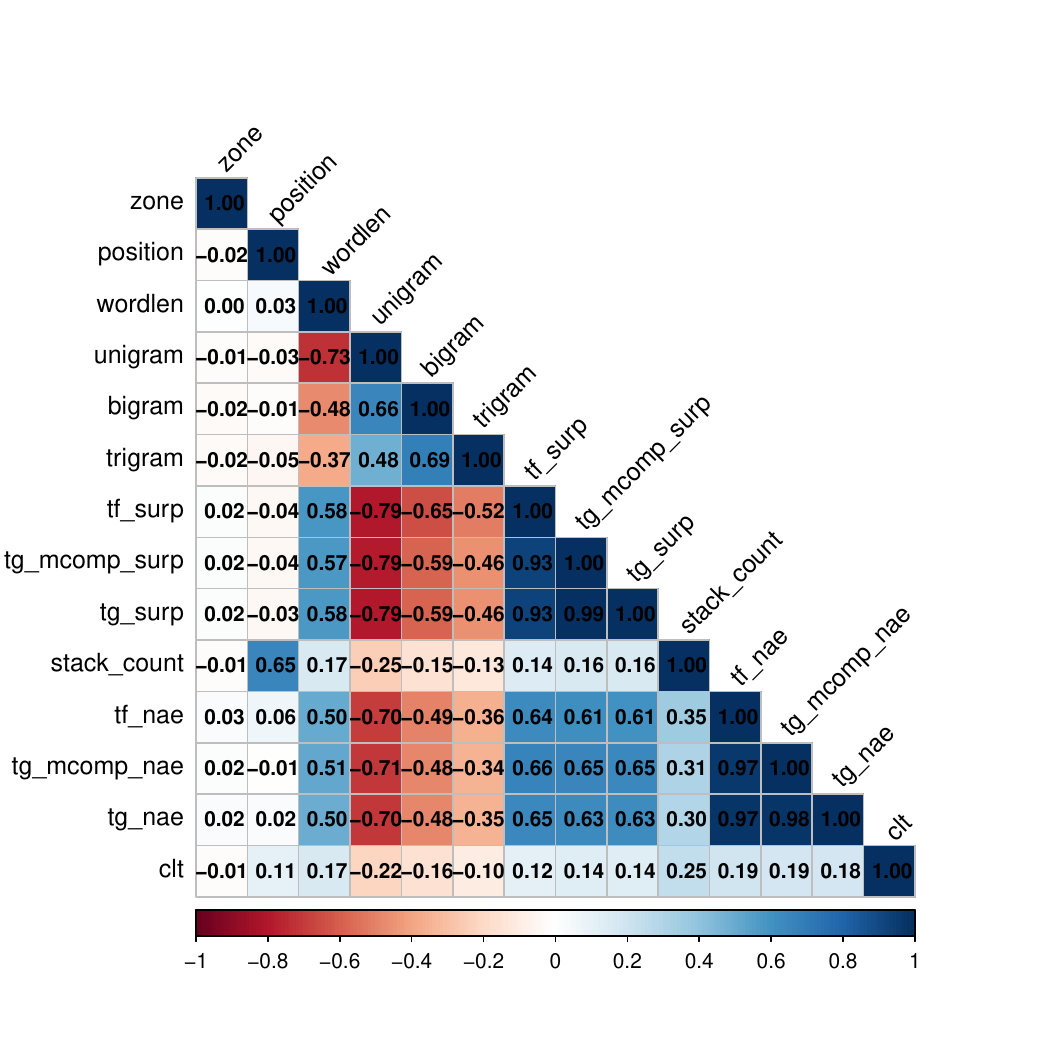}
    \caption{Correlations between the predictors in our statistical analysis}
    \label{fig:corr}
\end{figure*}

The correlations between the predictors in our statistical analysis are shown in Table~\ref{fig:corr}. While the NAE from different LMs shows very high correlations with each other, TG and Transformer provide independent predictive power for the self-paced reading times; TG subsumes the predictive power of TG\textsubscript{\comp} (see Section~\ref{subsec:nested} and~\ref{subsec:compose}).\footnote{\texttt{\_mcomp} indicates \comp.}

\section{Part-of-speech tags}
\label{app:pos}

Table~\ref{tab:pos} presents the complete list of part-of-speech (POS) and symbol tags in the Natural Stories corpus. As reading times are annotated for each whitespace-delimited region, for data points containing symbol tags (e.g., \texttt{NNP.}), we used the stripped version (e.g., \texttt{NNP}) in our analysis. Additionally, we excluded from our analysis any data points containing multiple POS tags (e.g., \texttt{NNP POS}).

\begin{table*}
    \centering
    \begin{tabular}{ll@{\hspace{2em}}ll}
        \hline
        \texttt{CC} & Coordinating conjunction & \texttt{PRP\$} & Possessive pronoun \\
        \texttt{CD} & Cardinal number & \texttt{RB} & Adverb \\
        \texttt{DT} & Determiner & \texttt{RBR} & Adverb, comparative \\
        \texttt{EX} & Existential \textit{there} & \texttt{RBS} & Adverb, superlative \\
        \texttt{FW} & Foreign word & \texttt{RP} & Particle \\
        \texttt{IN} & Preposition or subordinating conjunction & \texttt{TO} & \textit{to} \\
        \texttt{JJ} & Adjective & \texttt{UH} & Interjection \\
        \texttt{JJR} & Adjective, comparative & \texttt{VB} & Verb, base form \\
        \texttt{JJS} & Adjective, superlative & \texttt{VBD} & Verb, past tense \\
        \texttt{MD} & Modal & \texttt{VBG} & Verb, gerund or present participle \\
        \texttt{NN} & Noun, singular or mass & \texttt{VBN} & Verb, past participle \\
        \texttt{NNS} & Noun, plural & \texttt{VBP} & Verb, non-3rd person singular present \\
        \texttt{NNP} & Proper noun, singuler & \texttt{VBZ} & Verb, 3rd person singular present \\
        \texttt{NNPS} & Proper noun, plural & \texttt{WDT} & Wh-determiner \\
        \texttt{PDT} & Predeterminer & \texttt{WP} & Wh-pronoun \\
        \texttt{POS} & Possessive ending & \texttt{WP\$} & Possessive wh-pronoun \\
        \texttt{PRP} & Personal pronoun & \texttt{WRB} & Wh-adverb \\
        \hline
        \texttt{-LRB-} & Left round bracket & \texttt{,} & Comma \\
        \texttt{-RRB-} & Right round bracket & \texttt{.} & Period \\
        \texttt{``} & Open double quotes & \texttt{:} & Colon \\
        \texttt{''} & Closing double quotes & & \\
        \hline
    \end{tabular}
    \caption{POS and symbol tags in the Natural Stories corpus}
    \label{tab:pos}
\end{table*}

\section{Parallel parsing experiment}
\label{app:beam}

\begin{table*}[]
    \centering
    \begin{tabular}{lllccc}
        \toprule
        Model & $\Delta$LogLik ($\uparrow$) & Predictor & Effect size [\si{\milli\second}] & $p$-value range & Significant seeds \\
        \midrule
        \multirow{2}{*}{TG} & \multirow{2}{*}{\textbf{56.0} ($\pm$5.5)} & \texttt{tg\_bs\_nae} & 1.08 ($\pm$ 0.1) & $<$\textbf{0.001} & 3/3 \\
        & & \texttt{tg\_bs\_nae\_so} & 2.01 ($\pm$ 0.1) & $<$\textbf{0.001} & 3/3 \\
        \midrule
        \multirow{2}{*}{Transformer} & \multirow{2}{*}{26.1 ($\pm$8.1)} & \texttt{tf\_nae} & 0.95 ($\pm$ 0.1) & $<$\textbf{0.001} & 3/3 \\
        & & \texttt{tf\_nae\_so} & 1.25 ($\pm$ 0.2) & $<$\textbf{0.001} & 3/3 \\
        \bottomrule
    \end{tabular}
    \caption{TG's and Transformer's NAE contribution to reading time prediction, where TG's NAE was calculated with multiple syntactic structures generated by word-synchronous beam search with RNNG}
    \label{tab:effect_size_bs}
\end{table*}

As a conceptual case study for the local ambiguity resolution in syntactic structures behind token sequences, we implemented TG's NAE calculation using 10 syntactic structures obtained through word-synchronous beam search~\citep{stern2017Effective} with Recurrent Neural Network Grammar~(RNNG;~\citealp{dyer2016Recurrent,kuncoro2017What,noji2021Effective}).\footnote{\url{https://github.com/aistairc/rnng-pytorch}}\footnote{RNNG was trained on BLLIP-\textsc{lg} using default hyperparameters. For inference, action beam size and fast track were set to 100 and 1, respectively.} NAE was computed individually for each syntactic structure and then aggregated as a weighted average:
\begin{equation}
    \mathrm{NAE\_BS}_{l,h,w} \coloneqq \frac{\sum_{t \in \mathrm{Beam}_w} p(t) \cdot \mathrm{NAE}_{l,h,\tau(w,t)}^t}{\sum_{t \in \mathrm{Beam}_w} p(t)},
\end{equation}
where $\mathrm{Beam}_w$ represents the set of syntactic structures synchronized at the $w$-th word ($\lvert \mathrm{Beam}_w \lvert = 10$), and $\tau(w,t)$ denotes the token position corresponding to the $w$-th word in structure $t$.\footnote{\texttt{stack\_count} was similarly calculated as the weighted average across syntactic structures in $\mathrm{Beam}_w$.}

The analysis revealed patterns consistent with those observed when considering only the globally correct syntactic structure: both LMs' NAE demonstrated significant predictive power for reading times, with TG's NAE showing stronger contributions compared to Transformer's (Table~\ref{tab:effect_size_bs}).\footnote{\texttt{\_bs} indicates beam search.} The likelihood ratio tests further confirmed independent contributions from both LMs ($p<0.001$ for both comparisons: `TG~\&~Transformer'$>$`Transformer' and `TG~\&~Transformer'$>$`TG').

\section{Surprisal experiment}
\label{app:surp}

\begin{table*}[]
    \centering
    \begin{tabular}{lllccc}
        \toprule
        Model & $\Delta$LogLik ($\uparrow$) & Predictor & Effect size [\si{\milli\second}] & $p$-value range & Significant seeds \\
        \midrule
        \multirow{2}{*}{TG} & \multirow{2}{*}{62.6 ($\pm$6.8)} & \texttt{tg\_surp} & 1.36 ($\pm$ 0.0) & $<$\textbf{0.001} & 3/3 \\
        & & \texttt{tg\_surp\_so} & 1.88 ($\pm$ 0.1) & $<$\textbf{0.001} & 3/3 \\
        \midrule
        \multirow{2}{*}{Transformer} & \multirow{2}{*}{\textbf{159} \phantom{.}($\pm$12)} & \texttt{tf\_surp} & 2.47 ($\pm$ 0.1) & $<$\textbf{0.001} & 3/3 \\
        & & \texttt{tf\_surp\_so} & 2.98 ($\pm$ 0.2) & $<$\textbf{0.001} & 3/3 \\
        \bottomrule
    \end{tabular}
    \caption{TG's and Transformer's surprisal contribution to reading time prediction}
    \label{tab:effect_size_surp}
\end{table*}

We analyzed each LM's surprisal contribution to reading time prediction using a baseline regression model that excluded both LMs' surprisal from Equation~\ref{eq:regression} but included their NAE (Table~\ref{tab:effect_size_surp}). While both LMs' surprisal demonstrated significant predictive power for reading times, Transformer's surprisal exhibited a stronger contribution compared to TG's. Additionally, our likelihood ratio tests using the averaged surprisal revealed that the regression model incorporating both LMs' surprisal showed significantly higher predictive power compared to models with either LM's surprisal alone ($p<0.001$ for both comparisons: `TG~\&~Transformer'$>$`Transformer' and `TG~\&~Transformer'$>$`TG'). These findings suggest that (i) unlike attention mechanisms, next-token prediction based solely on token sequences more effectively captures dominant factors of human predictive processing, but (ii) similar to attention mechanisms, both types of next-token prediction---those based on token sequences alone and those leveraging both syntactic structures and token sequences---may coexist as models that capture distinct aspects of human predictive processing.

\section{Direct comparison of TG and TG\texorpdfstring{\textsubscript{\comp}}{comp}}
\label{app:weak}

\begin{table*}[]
    \centering
    \begin{tabular}{lllccc}
        \toprule
        Model & $\Delta$LogLik ($\uparrow$) & Predictor & Effect size [\si{\milli\second}] & $p$-value range & Significant seeds \\
        \midrule
        \multirow{2}{*}{TG} & \multirow{2}{*}{\textbf{78.2} ($\pm$6.9)} & \texttt{tg\_nae} & 1.43 ($\pm$ 0.2) & $<$\textbf{0.001} & 3/3 \\
        & & \texttt{tg\_nae\_so} & 2.28 ($\pm$ 0.3) & $<$\textbf{0.001} & 3/3 \\
        \midrule
        \multirow{2}{*}{TG\textsubscript{\comp}} & \multirow{2}{*}{59.4 ($\pm$17)} & \texttt{tg\_mcomp\_nae} & 1.36 ($\pm$ 0.2) & $<$\textbf{0.001} & 3/3 \\
        & & \texttt{tg\_mcomp\_nae\_so} & 1.90 ($\pm$ 0.2) & $<$\textbf{0.001} & 3/3 \\
        \bottomrule
    \end{tabular}
    \caption{TG's and TG\textsubscript{\comp}'s NAE contribution to reading time prediction with Transformer's NAE excluded from the regression baseline model}
    \label{tab:effect_size_weak}
\end{table*}

In Section~\ref{subsec:compose}, we explored the advantage of treating closed phrases as single representations beyond syntactic structure consideration. Our analysis incorporated Transformer's NAE in the baseline regression model to distinguish between two effects in TG\textsubscript{\comp}: syntactic structure consideration and direct terminal token access.

To evaluate which model---TG or TG\textsubscript{\comp}---better captures more dominant factors in human sentence processing as a single model, we assessed their predictive power without Transformer's NAE in the baseline regression model (Table~\ref{tab:effect_size_weak}). The analysis revealed TG's superior predictive power ($\Delta$LogLik=78.2) compared to TG\textsubscript{\comp} ($\Delta$LogLik=59.4). These results highlight that TG, which explicitly treats closed phrases as single representations, outperforms TG\textsubscript{\comp}, even when considering TG\textsubscript{\comp}'s advantage in direct terminal token access. The likelihood ratio tests confirmed TG's independent predictive power from TG\textsubscript{\comp} (`TG~\&~TG\textsubscript{\comp}'$>$`TG\textsubscript{\comp}', $p<0.001$), while in contrast to Section~\ref{subsec:compose}, TG\textsubscript{\comp} also accounted for unique variance (`TG~\&~TG\textsubscript{\comp}'$>$`TG', $p<0.01$). This bidirectional relationship likely emerges because TG\textsubscript{\comp}'s direct terminal token access explains unique variance in the absence of Transformer's NAE.

\section{License}

\begin{table*}[]
    \centering
    \begin{tabular}{ll}
        \toprule
        Dataset/Tool & License \\
        \midrule
        BLLIP~\citep{charniakeugene2000BLLIP} & BLLIP 1987--89 WSJ Corpus Release 1 \\
        Natural Stories corpus~\citep{futrell2018Natural} & CC BY-NC-SA 4.0 \\
        \midrule
        \texttt{transformer\_grammar}~\citep{sartran2022Transformera} & Apache 2.0 \\
        \texttt{rnng-pytorch}~\citep{noji2021Effective} & MIT License \\
        SentencePiece~\citep{kudo2018SentencePiece} & Apache 2.0 \\
        R (version 4.4.2)~\citep{rcoreteam2024Language} & GNU GPL $\ge$ 2 \\
        \texttt{lme4} (version 1.1.34)~\citep{bates2015Fitting} & GNU GPL $\ge$ 2 \\
        \texttt{lmerTest} (version 3.1.3)~\citep{kuznetsova2017LmerTest} & GNU GPL $\ge$ 2 \\
        \bottomrule
    \end{tabular}
    \caption{Licenses of datasets and tools}
    \label{tab:license}
\end{table*}

Table~\ref{tab:license} summarizes the licenses of the data and tools employed in this paper. All data and tools
were used under their respective license terms.

\end{document}